\documentclass{article}
\usepackage{spconf,amsmath,graphicx,epsfig,url}

\usepackage{color}
\usepackage{multirow}
\usepackage{booktabs}
\usepackage{array}
\usepackage{booktabs}
\usepackage{amssymb}
\usepackage{makecell}
\usepackage{cleveref}
\newcommand\etal{{\it et~al.~}}

\title{UMAAF: Unveiling Aesthetics via Multifarious Attributes of Images}
\name{Weijie Li$^{1, \star}$, Yitian Wan$^{1, \star}$, Xingjiao Wu$^{2, \dagger}$, Junjie Xu$^{1}$, Cheng Jin$^{2}$, Liang He$^{1}$
\thanks{$\star$ Weijie Li andYitian Wan contributed equally to this work. $\dagger$ Corresponding author: Xingjiao Wu (e-mail: xjwu\_cs@fudan.edu.cn).}
}
\address{$^{1}$East China Normal University, Shanghai, China\\
$^{2}$Fudan University, Shanghai, China
\\ \
}

\begin{document}

\UseRawInputEncoding
\maketitle

\begin{abstract}

With the increasing prevalence of smartphones and websites, Image Aesthetic Assessment (IAA) has become increasingly crucial. 
While the significance of attributes in IAA is widely recognized, many attribute-based methods lack consideration for the selection and utilization of aesthetic attributes. Our initial step involves the acquisition of aesthetic attributes from both intra- and inter-perspectives. Within the intra-perspective, we extract the direct visual attributes of images, constituting the absolute attribute. In the inter-perspective, our focus lies in modeling the relative score relationships between images within the same sequence, forming the relative attribute.
Then, to better utilize image attributes in aesthetic assessment, we propose the \textbf{U}nified \textbf{M}ulti-attribute \textbf{A}esthetic \textbf{A}ssessment \textbf{F}ramework (UMAAF) to model both absolute and relative attributes of images. 
For absolute attributes, we leverage multiple absolute-attribute perception modules and a absolute-attribute interacting network. 
The absolute-attribute perception modules are first pre-trained on several absolute-attribute learning tasks and then used to extract corresponding absolute attribute features. 
The absolute-attribute interacting network adaptively learns the weight of diverse absolute-attribute features, effectively integrating them with generic aesthetic features from various absolute-attribute perspectives and generating the aesthetic prediction.
To model the relative attribute of images, we consider the relative ranking and relative distance relationships between images in a Relative-Relation Loss function, which boosts the robustness of the UMAAF.
Furthermore, UMAAF achieves the state-of-the-art performance on TAD66K and AVA datasets, and multiple experiments demonstrate the effectiveness of each module and the model's alignment with human preference.

\end{abstract}
%
% \begin{keywords}
% Crowd counting, adaptive scenario discovery, convolutional neural network.
% \end{keywords}

\section{Introduction}
\label{sec:intro}

\begin{figure}[h]
   \centering
   \includegraphics[width=1\linewidth,height=4.5
   cm]{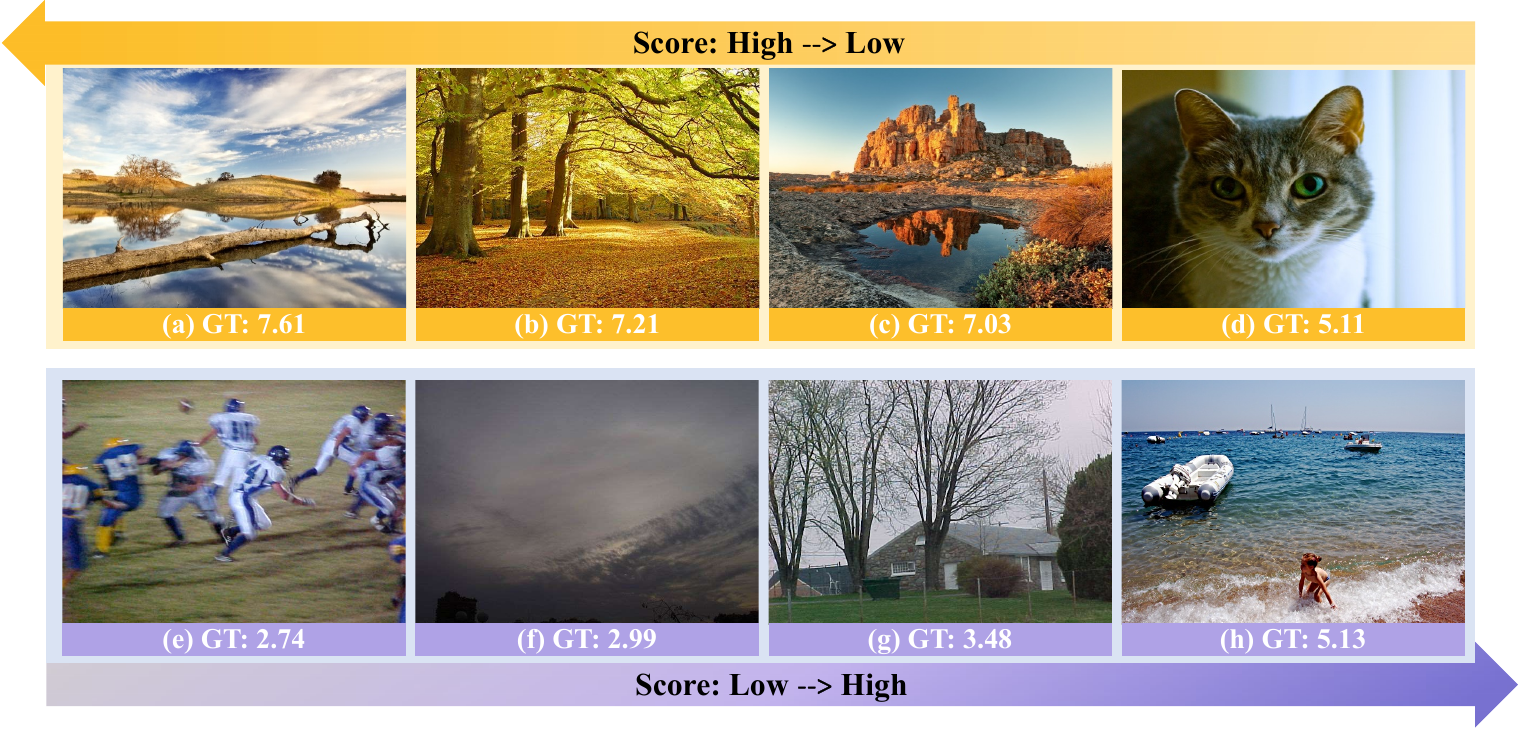}
   % \caption{Images with different aesthetic scores and `GT' means ground truth. The visual elements can directly influence their aesthetic scores and the relation between them are also important for both human and models' aesthetic assessment.}
  \caption{Images with diverse aesthetic scores (1-10) and `GT' denotes ground truth. 
The first row's image scores are ranked high to low from left to right. Evaluators typically rate \cref{fig:fig1}(a) - \cref{fig:fig1}(c) first and then \cref{fig:fig1}(d), often resulting in a lower score due to its aesthetic difference compared to the other three images. The second row follows an opposing trend.
}
   \label{fig:fig1}
    \vspace{-8pt}
\end{figure}

Considering the incredibly ever-growing scale of visual data spreading online, Image Aesthetic Assessment (IAA) appears to be particularly significant for numerous downstream applications, such as image retrieval \cite{9312626}, photography \cite{7456258}, and image search \cite{7974874}. Current research on aesthetic and psychology \cite{dewey2008art} have proved that any form of aesthetic experience comes down to a specific nervous system. Other studies \cite{li2023theme,he2022rethinking,cui2018distribution} attempt to simulate the human aesthetic assessment process by modeling aesthetic attributes of the image itself have got remarkable results. 

Although previous studies have demonstrated the efficacy of image attributes in facilitating the completion of IAA tasks is crucial, there are still two limitations. On the one hand, for attributes extracted from the image itself, they exhibit a lack of comprehensiveness \cite{cui2018distribution, he2022rethinking,li2023theme, celona2022composition}. For example, the attribute like `rule-of-third' is a common attribute used in previous studies, but it is only a part of the image composition. Besides, there is still room for exploration in the utilization of these extracted attributes. On the other hand, psychological research \cite{ariely2008predictably} suggests that the relative comparison between images can affect the results of aesthetic assessment, which is ignored by previous studies. 

To alleviate the aforementioned challenges, we propose to extract the aesthetic attributes from intra- and inter-perspectives of images. 
For the attributes extracted from the intra-perspective of images, specific elements such as composition, theme, and color play a direct and significant role in shaping the final aesthetic judgment made by individuals. For example, as illustrated in \cref{fig:fig1} (c), the composition of an image contributes to the perceived spatial relationship between the primary subject and the background. When combined with other inherent attributes of the image, this composition leads to a visually pleasing experience, resulting in a high aesthetic score. These essential components, exerting a direct impact on the visual experience of individuals, are denoted as `absolute attributes'.

For the attributes extracted from the inter-perspective of images, they mainly come from the comparison of the images' aesthetics within the same sequence, which has been verified by psychological research \cite{ariely2008predictably} that they can influence people's aesthetic judgments, leading to potentially cognitive biases. For example, the evaluation of images within two rows where the assessment progresses from left to right. In the case of assessing \cref{fig:fig1} (d) and \cref{fig:fig1} (h), the final images in their respective rows, evaluators tend to assign lower scores to \cref{fig:fig1} (d) compared to a direct evaluation, while the opposite tends to happen for \cref{fig:fig1} (h). This phenomenon underscores the importance of considering the order in which images are presented.
In the context of models' aesthetic assessment, a notable example occurs when the training data lacks shuffling. This oversight can result in models learning incorrect relational features, impairing their ability to accurately assess aesthetics. This type of factor, which relies on comparisons or the arrangement of data rather than being directly derived from individual images and indirectly influences human visual judgments of images, is indicated as the `relative attribute'.

To harness the complete potential of aesthetic evaluation, incorporating both the absolute and relative attributes of images, we introduce the Unified Multi-Attribute Aesthetic Assessment Framework (UMAAF). This innovative framework integrates novel models and advanced loss functions to enhance the overall assessment process.

To address the absolute attributes, we introduce a purpose-built network aimed at their extraction and fusion. 
Guided by classical principles in photography \cite{barnbaum2017art}, we pinpoint four fundamental and comprehensive attributes: Composition, Color, Exposure, and Theme, forming the foundation for our novel feature extraction component.
Distinct from conventional direct feature concatenation methods \cite{he2022rethinking,cui2018distribution}, we introduce a absolute-attribute interacting  network. This component effectively merges the features extracted from absolute attributes with the overall aesthetic feature obtained from a shared network. To achieve this, we fuse the features from attributes perspectives and leverage bilinear fusion technique \cite{Mao_Zhu_Su_Cai_Li_Dong_2023} within the fusion module, ultimately generating an aesthetic prediction score. 
In the realm of relative attributes, there has been a noticeable scarcity of IAA studies with a dedicated focus on this aspect. To address the critical need for modeling the relative relationships among images, we devise an innovative loss function named the Relative-Relation Loss, which is implemented within the framework of triplet loss \cite{schroff2015facenet}.
The Relative-Relation Loss captures the intricate interplay between images by considering both their relative rankings and the distance relationships. 

In a nutshell, our main contributions are summarized as follows:
\begin{itemize}
\item \textbf{Unified Multi-Attribute Aesthetic Assessment Framework (UMAAF):} We introduce a comprehensive framework that takes both absolute and relative attributes of images into account. This framework incorporates multiple components, where each component is designed to handle a specific aspect of the image aesthetics. Additionally, we propose a novel loss function to capture the relative attribute information, enabling a more comprehensive perspective in learning the IAA task.

\item \textbf{Efficient Extraction of Absolute Attributes:} We efficiently extract concrete absolute-attribute features from images, following real photographic rules. This is achieved through the deployment of multiple Absolute-Attribute Perception Components, where each component is designed to capture a specific absolute attribute. We ensure a comprehensive integration of these extracted features through our Absolute-Attribute Interacting Network, providing a holistic view.

\item \textbf{Modeling Relative Attribute Information:} To effectively capture the relative attribute information between images, we introduce the Relative-Relation Loss. This loss function takes both the relative rankings and distance relationships between images into consideration, enhancing the model's ability to perceive the aesthetic differences between images.

\item \textbf{Empirical Performance and Alignment with Human Preference: }Through extensive experiments, we showcase that our model not only achieves state-of-the-art performance on aesthetic datasets but also closely aligns with human preference. Furthermore, our additional experiments validate the effectiveness of each component within our proposed model.
\end{itemize}

\section{Related Work}
\label{sec:related}
\begin{figure*}[t]
   \centering
    \includegraphics[width=0.98\linewidth]{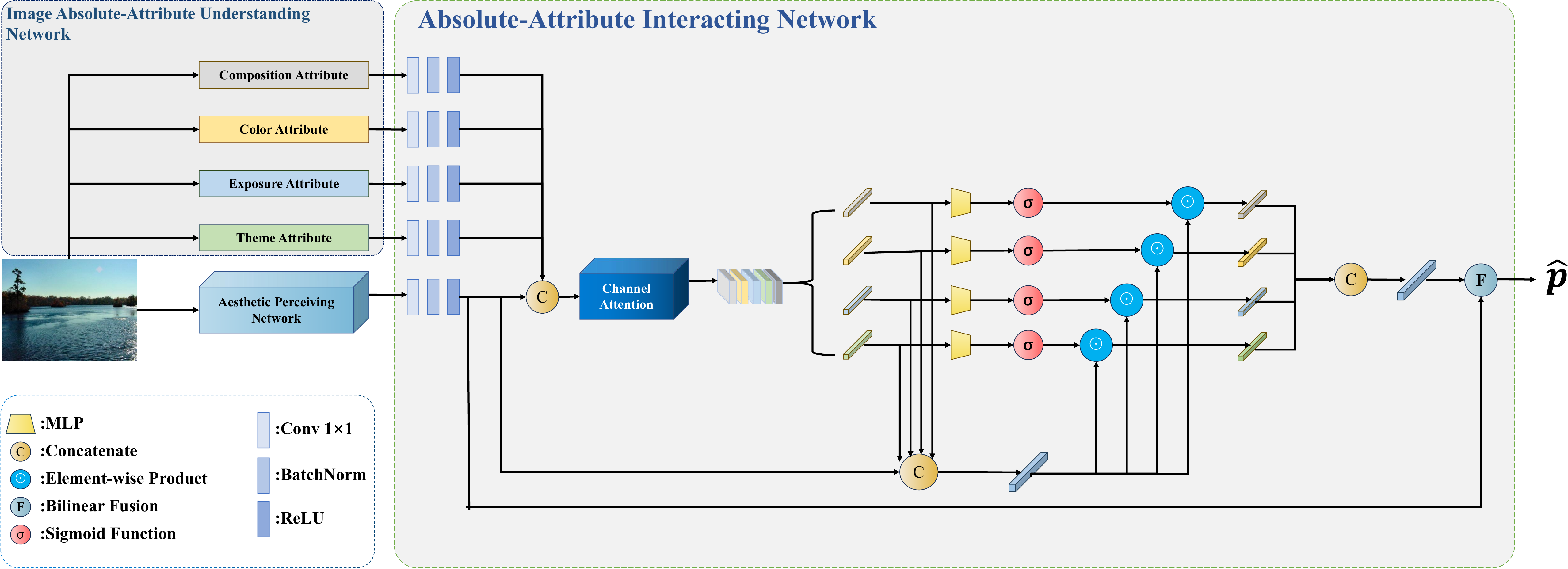}
    \caption{Overview of the UMAAF. The whole architecture can be divided into three components: Image Absolute-Attribute Understanding Network, Absolute-Attribute Interacting Network and Aesthetic Perceiving Network. Image Absolute-Attribute Understanding Network has four branches used to extract corresponding attribute features, Absolute-Attribute Interacting Network adaptively fuses different features and Aesthetic Perceiving Network is mainly a MobileNetV2 network used to extract generic aesthetic features. }
%\caption{Overview of the UMAAF. }
   \label{fig:overview}
\vspace{-8pt}
\end{figure*}

\subsection{Image Aesthetic Assessment Methods}

IAA has seen decades of evolution. Initially, hand-crafted features were used to represent image aesthetics \cite{obrador2010role,datta2006studying,nishiyama2011aesthetic,sun2009photo}, but their limited capacity to handle diverse and complex images became evident. The emergence of deep learning revolutionized IAA, with researchers shifting to CNN or DNN-based approaches to learn aesthetic features, progressively replacing handcrafted ones.
Early attempt like the RAPID model \cite{lu2014rapid}, was trained on IAA datasets, followed by methods such as DMA-Net \cite{lu2015deep} that used multiple patches in image training and A-Lamp \cite{ma2017lamp} with a two-stream architecture for global and local features. NIMA \cite{talebi2018nima} introduced an Earth Mover's Distance loss for more accurate aesthetic score predictions, offering a distribution of scores rather than the simple binary classification.

A significant advancement is the introduction of the multi-level spatially pooled (MLSP) feature \cite{hosu2019effective}, derived from InceptionNet \cite{inceptionresnetv2} convolution blocks, yielding more comprehensive image aesthetic representations. She \etal. \cite{she2021hierarchical} used a multi-layer graph neural network to extract image composition information and achieves the best results in aesthetic classification tasks. Hou \etal. \cite{hou2020object} utilized object detection techniques to identify multiple objects within images, and then evaluate the image aesthetic based on the relation between objects. Pre-training strategies, such as adapting networks from image editing tasks \cite{sheng2020revisiting} or expanding editing operations \cite{yi2023towards}, led to notable improvements.
However, these methods primarily focus on visual image features and often overlook the direct influence of image attributes on human perception. This aspect, which plays a crucial role in shaping human feelings toward images, is often neglected in these approaches.

\subsection{Attribute-aware Aesthetic Assessment}

As the importance of attributes in shaping human perception of image aesthetics became evident, attribute-aware approaches emerged. The SANE \cite{cui2018distribution} employed pre-trained deep networks to detect objects and extract scene information, which is then combined with aesthetic features. Celona \etal. \cite{celona2022composition} focused on abstract attributes like composition and style, utilizing a hypernet to assess image aesthetics based on these extracted attributes. Adversarial learning is used in \cite{pan2019image} to incorporate attributes like lights and the rule of thirds, and a multi-task deep network predicts both aesthetic scores and attributes, distinguished by a discriminator. Li \etal. \cite{li2023theme} involved pre-training tasks for multiple attributes, then using graph neural network to fuse different features. He \etal \cite{he2023thinking} delved into the significance of color in images for their aesthetic appeal. Notably, TANet \cite{he2022rethinking} emphasized theme attributes, using a dedicated network to extract different theme rules from diverse images.

Compared to prior attribute-aware methods, our approach advances by extracting and modeling absolute and relative attributes simultaneously. For absolute attributes, our selection is based on pragmatic and comprehensive considerations. Furthermore, diverging from previous methodologies that placed lesser emphasis on integrating attribute features, we aim to investigate innovative techniques for combining these features. In the realm of relative attributes, a previously unexplored dimension, we introduce a loss function to capture inter-image relative relationships. This enriches the model's understanding of aesthetic considerations within specific contextual settings.

\section{Framework}
In this section, we will start by introducing our overall framework. Then, we will separately present the Image Absolute-Attribute Understanding Network and the Absolute-Attribute Interacting Network. Finally, we will provide a detailed explication of the Relative-Relation Loss.

\begin{figure*}[t]
\centering
\includegraphics[width=0.98\linewidth]{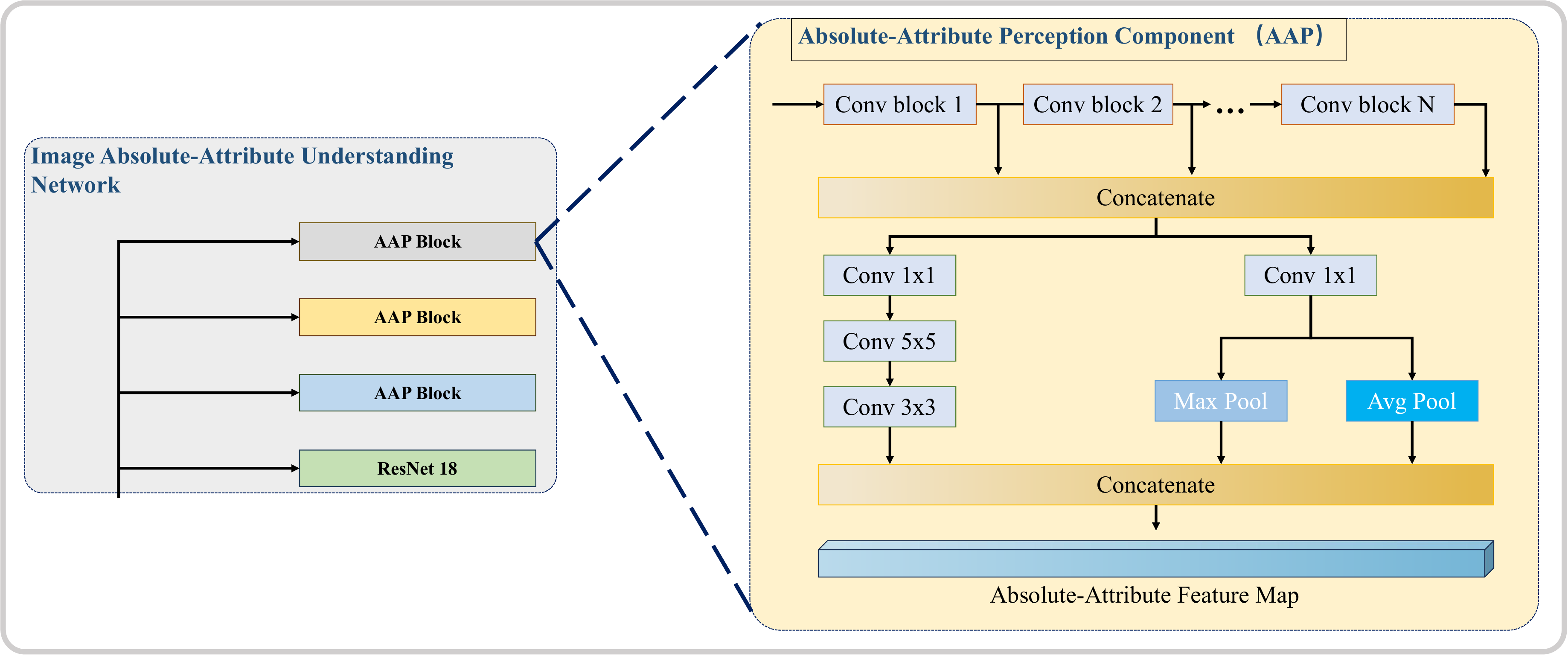}
%\caption{The architecture of Absolute-Attribute Perception Component. Two streams are used to further concentrate the long-range information and retained significant information separately.}
\caption{The architecture of Absolute-Attribute Perception Component.}
\label{fig:attribute perception module}
\vspace{-8pt}
\end{figure*}

\subsection{Method}
The architecture of the UMAAF is shown in \cref{fig:overview}. The network architecture is specifically divided into three modules: 1) Image Absolute-Attribute Understanding Network: It extracts features that are relevant to determine absolute attributes from the image. 2) Aesthetic Perceiving Network: It mainly consists of a MobileNetV2 \cite{mobilenetv2} and generates the generic aesthetic feature. 3) Absolute-Attribute Interacting Network: It interacts with the feature of each attribute and outputs the aesthetic prediction.

\subsection{Absolute-Attribute Understanding Network}
Inspired by classical rules of photography \cite{barnbaum2017art}, we begin by identifying the absolute attributes to be perceived: Composition, Color, Exposure, and Theme. For the first three attributes, we employ three pre-trained Absolute-Attribute Perception Components, while the theme features are extracted by using a mature method from previous works \cite{he2022rethinking, cui2018distribution, li2023theme}.

The architecture of the Absolute-Attribute Perception Component is illustrated in \cref{fig:attribute perception module}. Following the approach from \cite{hosu2019effective}, we use InceptionResNet-v2 \cite{inceptionresnetv2} as the backbone, leveraging the multi-level features of the image. The component is divided into two parts: one utilizes a stack of convolution layers with various kernel sizes to focus on long-range information in image features through a larger receptive field, while the other part diversely aggregates and retains information using both the Average Pooling layer and Max Pooling layer.

We subsequently apply corresponding output layers and loss functions based on the specific pre-trained tasks. Next, we introduce the pre-training tasks of each absolute-attribute extractor.

\noindent\textbf{Composition Attribute. }
Image composition quality is a crucial factor in assessing the aesthetic appeal of an image. We train our Attribute Perception Component on the CADB dataset \cite{zhang2021image}, which serves as an image composition attribute extractor. The CADB dataset, comprising 9,958 real-world images, provides composition quality scores ranging from 1 to 5, where higher scores indicate better composition quality.
It's noteworthy that our attribute perception component, trained on the official partition of CADB, achieves remarkable results on its test set, with a PLCC (Pearson Linear Correlation Coefficient) of 0.715 and a SRCC (Spearman Rank Correlation Coefficient) of 0.705. These scores significantly outperform the state-of-the-art method on CADB as reported in the original paper \cite{zhang2021image} (PLCC: 0.671, SRCC: 0.656). This underscores the effectiveness of our Absolute-Attribute Perception Component in extracting image attributes, particularly in the context of image composition, validating the efficiency of our composition attribute extraction component.

\noindent\textbf{Color Attribute. }
The image's color significantly impacts its visual appeal. Vibrant colors often enrich the visual experience, while subdued colors with unique compositions or themes can evoke distinct emotions. To quantify this, we adopt the colorfulness classification method from \cite{measurecolor2003}, categorizing each image in the AVA dataset into 7 colorfulness levels: not colorful, slightly colorful, moderately colorful, averagely colorful, quite colorful, highly colorful, and extremely colorful.
Besides, we pre-train an Absolute-Attribute Perception Component for the image color richness classification task. This enables the component to focus on regions of the image related to color, ultimately yielding a color feature extraction component.

\noindent\textbf{Exposure Attribute. }
The exposure level of a photo, as highlighted in \cite{afifi2021learning}, significantly impacts its overall visual experience. Overexposure can make an image too bright, while underexposure can render it too dark, leading to distinct lighting conditions. We employ the dataset from \cite{afifi2021learning} for pre-training, categorizing images into 5 exposure value (EV) categories.
By conducting this pre-training task to classify exposure values, our model effectively focuses on the variations in light and shadow within the image. Consequently, we obtain an exposure and light feature extraction component.

\noindent\textbf{Theme Attribute. }
Recent studies \cite{he2022rethinking,li2023theme} highlight the significant correlation between the theme of an image and its overall aesthetics. Inspired by the approach in TANet \cite{he2022rethinking}, we employ a pre-trained ResNet18 \cite{resnet} model, trained on the Places dataset \cite{place365}, as the image theme attribute extractor. The Places dataset comprises approximately 10 million images, covering over 400 distinct scene label categories. We utilize this branch to more effectively extract theme-related information from the image.

\subsection{Absolute-Attribute Interacting Network}
Most current IAA methods lack in-depth exploration of attribute fusion techniques. To address this, we introduce a Absolute-Attribute Interacting Network to fully utilize various attribute features.

Recognizing that different absolute attributes hold varying levels of importance for the final result on a single image, we employ the channel attention mechanism \cite{cbam} to effectively weight these diverse absolute-attribute features. The input to the attention component is a concatenation of the aesthetic features and all absolute-attribute features, allowing us to leverage the image's features fully. The attention component utilizes Max Pooling and Average Pooling to capture different context information and a shared parameter Multilayer Perceptron (MLP) to compute the ultimate attention weights.

After that, we get out the corresponding absolute-attribute features from the concatenation and transform them into feature vectors by an Average Pooling Layer. Then, inspired by \cite{Mao_Zhu_Su_Cai_Li_Dong_2023}, we use a feature selection mechanism to further integrate the absolute-attribute and aesthetic features from different attribute perspectives. Specifically, for each absolute attribute, we use a gate component to get attribute-specific features. The whole process can be described as follows:
\begin{equation}
\begin{aligned}
z_i &= \sigma(MLP_i(x_i)) \odot o , i\in\mathbf{S},
\end{aligned}
\end{equation}
where \textbf{S} denotes the absolute-attributes set, including the absolute attributes determined before, $x_i$ denotes the absolute-attribute feature vectors and $o$ denotes the concatenation of all features. We do element-wise products using weights and concatenation of all features to get multiple attribute-specific features $z_i$. 

Finally, we concatenate all attribute-specific features and use the bilinear fusion \cite{Mao_Zhu_Su_Cai_Li_Dong_2023} to integrate them with generic aesthetic features and get the prediction. 
 
\subsection{Relative-Relation Loss}
Current aesthetic evaluation methods \cite{talebi2018nima,icpr2022} pay more attention to modeling the features of the image itself, and the relation between images is rarely emphasized. Due to the importance of the relation of images in IAA, we develop Relative-Relation Loss to model the relative attributes of images and because of the feasibility, our loss function considers the samples in the same batch. As for the loss, first, consider i, j, k represent three samples in one batch, and their ground truth label is $g_i, g_j,g_k$, while meeting the condition that $g_i > g_j > g_k $ or $g_i < g_j < g_k$. We use sample i as the anchor and sample j, k as the positive and the negative one respectively. A triplet loss \cite{schroff2015facenet} is used to perform the following calculation:
\begin{equation}
\begin{aligned}
L_{trp}(i,j,k)=max\{0, |p_i-p_j|-|p_i-p_k|+|g_j-g_k|\},
\end{aligned}
\end{equation}
where $p_i, p_j, p_k$ are the predicted values of three samples.

The introduced triplet loss function aims to increase the gap between the predicted scores of the anchor sample and the negative sample, while decreasing the gap between the anchor sample and the positive sample's predicted scores. The objective is to align these predicted scores' distances more closely with the distances between the ground truth. As such, the margin for the triplet loss is established as $|g_j-g_k|$, calculated based on a hypothetically perfect prediction environment \cite{9706735}.

Based on the triplet loss, considering the relationship between the sample and the other samples in the same batch, the Relative-Relation Loss can be formulated as follows:

\begin{equation}
\begin{aligned}
L_{relative} &= \frac{1}{b-4}\sum_{i=3}^{b-2}(\frac{1}{b-3}
(\sum_{j=2}^{i-1}L_{trp}(i,j,j-1)\\+&\sum_{j=i+1}^{b-1}L_{trp}(i,j,j+1))
),
\end{aligned}
\end{equation}
where $b$ means the batch size. When training, we sort the samples in one batch by ground truth labels from the largest to the smallest firstly. 
Then, samples in the same batch are used as anchors in turn, and other neighboring samples are selected as positive samples and negative samples in pairs.

We utilize the loss as defined in ~\cref{Eqloss} to constrain the UMAAF:

\begin{equation}
\begin{aligned}
L_{total}=L(\hat{p},p)+\lambda L_{relative},
\end{aligned}
\label{Eqloss}
\end{equation}
where $\lambda$ is the balancing coefficient and set to 0.05 during training. $\hat{p}$ and $ p$ represent the output of the model and the label respectively. The L(i,j) is the Earth Mover's Distance(EMD) \cite{talebi2018nima} loss if the label is the aesthetic distribution or Mean Squared Error(MSE) loss when the label is aesthetic score.

\section{Experiments}
\label{sec:exp}

\begin{table*}[!t]
\centering
%\caption{Comparison of the proposed model with the state-of-the-art IAA methods on TAD66K and AVA. '-' means the results are not available in known papers. The results on TAD66K are mainly from \cite{he2022rethinking} and the results on AVA are from respective papers.}
% \tabcolsep=0.35cm
% %\label{tab:allresults}
% \renewcommand\arraystretch{1.2}
\caption{Comparison of the proposed model with the state-of-the-art IAA methods on TAD66K and AVA. '-' means the results are not available in known papers. The results on TAD66K are mainly from \cite{he2022rethinking} and the results on AVA are from respective papers.}
\begin{tabular}{p{68px}p{46px}<{\centering}p{46px}<{\centering}p{46px}<{\centering}|p{46px}<{\centering}p{46px}<{\centering}p{46px}<{\centering}p{46px}<{\centering}}
\toprule[1.0pt]
% \usepackage{booktabs}
\multirow{2}{*}{Method} &\multicolumn{3}{c|}{TAD66K} &\multicolumn{4}{c}{AVA} \\ \cline{2-4} \cline{5-8}  ~ &PLCC $\uparrow$ &SRCC $\uparrow$ &MSE $\downarrow$ & PLCC $\uparrow$  & SRCC $\uparrow$  & Accuracy $\uparrow$ & EMD $\downarrow$  \\  \midrule[1pt]
RAPID \cite{lu2014rapid}     & 0.332  & 0.314  & 0.022 & 0.453 & 0.447 & 71.18 & -     \\
ALamp \cite{ma2017lamp}      & 0.422  & 0.411  & 0.019 & 0.671 & 0.666 & 82.52 & -     \\
NIMA  \cite{talebi2018nima}  & 0.405  & 0.390  & 0.021 & 0.636 & 0.612 & 81.49 & 0.05  \\
MPada \cite{mpada}           & 0.480  & 0.466 & 0.022 & 0.731  & 0.727  & 83.03    & -     \\
MLSP \cite{hosu2019effective}& 0.508  & 0.490  & 0.019 & 0.757  & 0.756  & 81.72    & -     \\
UIAA \cite{Zeng2020AUP}      & 0.441  & 0.433 & 0.021 & 0.720  & 0.719  & 80.79    & 0.065 \\
GPF-CNN \cite{8691784}       & - & - & - & 0.704  & 0.690  & 81.81    & -     \\
AFDC \cite{9157018}          & - & - & - & 0.671  & 0.649  & 83.18    & -     \\
ReLIC \cite{ZHAO2020103024}  & - & - & - & 0.760  & 0.748  & 82.35    & -     \\ 
Xu \etal \cite{contextmm}            & - & - & - & 0.725  & 0.724  & 80.9     & -     \\
Hou \etal \cite{hou2020object}              & - & - & - & 0.753  & 0.751  & 81.67    & -     \\
HGCN  \cite{she2021hierarchical}         & 0.493  & 0.486 & 0.020 & 0.687  & 0.665  & \textbf{84.61}    & 0.043 \\
MUSIQ \cite{9710973}         & - & - & - & 0.738  & 0.726  & 81.5     & -     \\
TANet \cite{he2022rethinking}& 0.531  & 0.513 & 0.016 & 0.765  & 0.758  & 80.63    & 0.047 \\
GAT-GATP \cite{icpr2022}     & - & - & - & 0.764  & \textbf{0.762}  & -        & -     \\
TAVAR \cite{li2023theme}     & - & - & - & 0.736  & 0.725  & -        & -     \\\midrule[0.5pt]
\textbf{UMAAF}                         & \textbf{0.540}  & \textbf{0.515} & \textbf{0.015} & \textbf{0.770}  & 0.759  & 81.69    & \textbf{0.042}  \\ \bottomrule [1.0pt]
\end{tabular}
\label{tab:allresults}
\end{table*}

In this section, we will first describe the experimental setup for the pre-training phase and the overall training phase. Subsequently, we compare UMAAF with the state-of-the-art methods on the aesthetic datasets: AVA \cite{6247954} and TAD66K \cite{he2022rethinking}. Following the previous methods \cite{talebi2018nima,hosu2019effective}, we use PLCC, SRCC and Accuracy to evaluate the results. 
In addition, the MSE loss and EMD loss on test set \cite{he2022rethinking} are also included in the evaluation metrics of TAD66K\cite{he2022rethinking}  and AVA \cite{6247954} respectively. 
Finally, we perform ablation experiments to verify the validity of each component and selected attributes. 

\subsection{Implementation Details}
\textbf{Pre-training Settings. }
During the pre-training phase, we utilize a pre-trained InceptionResNet-v2 \cite{inceptionresnetv2} from ImageNet \cite{5206848} as the backbone for our Absolute-Attribute Perception Components. Each image used for pre-training is initially resized to $330 \times 330 \times 3$ and subsequently randomly cropped to $299 \times 299 \times 3$ for input. We employ an Adam optimizer with a batch size of 32 and apply a weight decay of $5e^{-4}$. The initial learning rate is set to $1e^{-4}$.

\noindent\textbf{Training Settings. }
During training, we keep the pre-trained feature extractors' parameters frozen. Images are resized to  $224\times224\times3$ for the backbone and theme attribute extractor and $299 \times299\times3$ for the other attribute extractors. Random horizontal flipping is used for data augmentation.
We utilize the Adam optimizer with a batch size of 32, weight decay of $5e^{-4}$, and an initial learning rate of $1e^{-6}$. If the network's loss doesn't decrease for 5 consecutive epochs, we reduce the learning rate by a factor of 0.1. Training stops when the loss ceases to decrease. The network is implemented in PyTorch and runs on an A100 GPU with 40GB memory.

\subsection{Performance Comparison}
The results of the proposed model and the comparison with other known methods are shown in \cref{tab:allresults}.

%\subsubsection{Performance on TAD66K Dataset. }
\noindent\textbf{Performance on TAD66K Dataset. } 
On the TAD66K dataset, UMAAF achieves state-of-the-art performance on all metrics. Compared with the previous work \cite{he2022rethinking}, UMAAF further models more absolute attributes in the network and different from directly concatenating multiple features, we further integrate various features from a more comprehensive perspective. The comparison results show the promotion from effective utilization of absolute attributes.

%\subsubsection{Performance on AVA Dataset. }
\noindent\textbf{Performance on AVA Dataset. } 
We outperform other methods in all metrics except SRCC and Accuracy. Among them, SRCC is close to the current best result and still competitive. Compared to previous works which mostly develop information within the images themselves, we additionally model the relative information between images and obtain a higher PLCC and significantly lower EMD value.

\begin{table}[t]
\centering
\caption{Comparison of the proposed model with several representative IAA methods on human preference.}
\begin{tabular}{p{80px}p{66px}<{\centering}p{66px}<{\centering}}
\toprule[1.0pt]
% \usepackage{booktabs}
Method                       & PLCC $\uparrow$  & SRCC $\uparrow$ \\ \midrule[1pt]
NIMA \cite{talebi2018nima}                        & 0.234  & 0.231 \\
MLSP \cite{hosu2019effective}                        & 0.246  & 0.241 \\
TANet \cite{he2022rethinking}                        & 0.249  & 0.250 \\ \midrule[1pt]
\textbf{UMAAF}                         & \textbf{0.261}  & \textbf{0.254}  \\ \bottomrule [1.0pt]
\end{tabular}
\label{tab:human}
% \vspace{-8pt}
\end{table}

\noindent\textbf{Performance on human preference. }
We employ ImageReward \cite{xu2023imagereward} to assess the model's compliance with human preference. 
ImageReward is used to evaluate the degree of human preference for text-to-image results. 
Giving a piece of text and an image to the model, it will return the score that reflects the image's degree of human preference. 
We use a text: ` a photograph of high aesthetic quality, with variable and real content ' to describe the images in the aesthetic dataset as the text input of ImageReward and combine it with the images in the dataset to get the model's assessment. We normalize the human preference score obtained by ImageReward and the aesthetic score obtained by the model, and then calculated the PLCC and SRCC between them to evaluate the correlation between the models' results and human preference. We selected NIMA \cite{talebi2018nima}, MLSP \cite{hosu2019effective}, TANet \cite{he2022rethinking} and UMAAF to compare, and the results are shown in \cref{tab:human}. The results prove that UMAAF is more consistent with human preference than other representative aesthetic assessment models.

\begin{table*}[t]
\centering
% \tiny
%\caption{Ablation study results of proposed method on TAD66K.}
% \tabcolsep=0.06cm
% \renewcommand\arraystretch{1.2}
%\label{tab:ablation}
\caption{Ablation study results on TAD66K.}
\begin{tabular}{p{36px}<{\centering}p{36px}<{\centering}|p{46px}<{\centering}p{62px}<{\centering}p{40px}<{\centering}p{40px}<{\centering}p{40px}<{\centering}p{40px}<{\centering}p{46px}<{\centering}}
\toprule[1.0pt]
% \usepackage{booktabs}
PLCC & SRCC & 
MobileNetV2 & 
\makecell{Composition \\ Attribute} & 
\makecell{Color \\ \ Attribute} & 
\makecell{Exposure \\  Attribute} &
\makecell{Theme \\ Attribute} &
\makecell{Relative \\  Loss} &
\makecell{Attribute \\ Interaction}  \\ \midrule[0.5pt]
0.456 & 0.447 & $\checkmark$ & & & & & & \\
0.471 & 0.453 & $\checkmark$ & & & & & $\checkmark$ & \\
0.485 & 0.475 & $\checkmark$ & $\checkmark$ & & & & & \\
0.473 & 0.467 & $\checkmark$ & & $\checkmark$ & & & & \\
0.478 & 0.477 & $\checkmark$ & & & $\checkmark$ & & & \\
0.481 & 0.471 & $\checkmark$ & & & & $\checkmark$ & & \\
0.505 & 0.482 & $\checkmark$ & $\checkmark$ & & & $\checkmark$ & &  \\
0.502 & 0.487 & $\checkmark$ & $\checkmark$ & $\checkmark$ & $\checkmark$ & & &\\
0.513 & 0.489 & $\checkmark$ & $\checkmark$ & $\checkmark$ & $\checkmark$ & $\checkmark$  & & \\
0.522 & 0.497 & $\checkmark$ & $\checkmark$ & $\checkmark$ & $\checkmark$ & $\checkmark$ & $\checkmark$ & \\
0.524 & 0.498 & $\checkmark$ & $\checkmark$ & $\checkmark$ & $\checkmark$ & $\checkmark$ & & $\checkmark$\\
\textbf{0.540} & \textbf{0.515} & $\checkmark$ & $\checkmark$ & $\checkmark$ & $\checkmark$ & $\checkmark$ & $\checkmark$ & $\checkmark$ \\
\bottomrule [1.0pt]
\end{tabular}
\label{tab:ablation}
% \vspace{-8pt}
%\caption{Ablation study results of proposed method on TAD66K.}
\end{table*}

\begin{table*}[t]
\centering
% \scriptsize
% \tabcolsep=0.1cm
% \renewcommand\arraystretch{1.2}
%\label{tab:att}
\caption{Results of different structures on extracting multi absolute attributes and the corresponding final results on TAD66K.}
\begin{tabular}{lp{56px}<{\centering}p{56px}<{\centering}|p{56px}<{\centering}|p{56px}<{\centering}|p{56px}<{\centering}|p{56px}<{\centering}}
\toprule[1.0pt]
\multirow{2}{*} {Models} &\multicolumn{2}{c|} {Composition} &\multicolumn{1}{c|} {Color}  &\multicolumn{1}{c|} {Exposure} &\multicolumn{2}{c} {TAD66K}  \\  \cline{2-3} \cline{4-4} \cline{5-5} \cline{6-7} ~ &  PLCC $\uparrow$ & SRCC $\uparrow$ & Accuracy $\uparrow$ & Accuracy $\uparrow$ & PLCC $\uparrow$ & SRCC $\uparrow$ \\ \midrule [0.5pt]
ResNet18    & 0.597         & 0.586       & 85.64$\%$         & 84.11$\%$             & 0.515          & 0.490   \\
ResNet34    & 0.613         & 0.603       & 86.60$\%$         & 84.96$\%$             & 0.518          & 0.492   \\
ResNet50    & 0.622         & 0.612       & 88.69$\%$         & 87.06$\%$             & 0.522          & 0.496        \\
InceptionResNetv2 & 0.642   & 0.634       & 88.98$\%$         & 88.34$\%$             & 0.522          & 0.501        \\
AAP (w/o cnn)  & 0.695    & 0.685    & 0.40$\%$            & 89.03$\%$                & 0.532        & 0.508                      \\
AAP (w/o pool) & 0.688    & 0.667    & 89.44$\%$            & 88.51$\%$               & 0.524        & 0.505                      \\
\textbf{AAP}          & \textbf{0.715}    & \textbf{0.705}    & \textbf{90.82\%}            &\textbf{89.84\%}              &  \textbf{0.540}        & \textbf{0.515}           \\ 
\bottomrule [1.0pt]
\end{tabular}
\label{tab:attribute structure}
% \vspace{-8pt}
\end{table*}

\subsection{Ablation Study}
\noindent\textbf{Effectiveness of modules and loss function. }
\cref{tab:ablation} summarizes the outcomes of our ablation study conducted on the TAD66K dataset \cite{he2022rethinking}.
Initially, we evaluate the impact of the loss function when applied solely to the backbone. The introduction of this loss function leads to a 3.3$\%$ increase in PLCC and a 1.4$\%$ increase in SRCC for the model.

Next, each absolute-attribute branch is individually added to the backbone to assess their effectiveness. Notably, the composition and theme branches exhibit the most significant improvements. Upon integrating the composition branches, PLCC and SRCC rise by 6.3$\%$ and 6.4$\%$ respectively, underscoring the importance of composition and theme in image aesthetics assessment. This reaffirms that absolute-attribute features aid the network in gaining a better understanding of image aesthetics.

Once all attributes are incorporated, we delve into the effects of the loss function and the Absolute-Attribute Interacting Network. When added to models that already possess all attribute branches, the proposed loss function results in a 1.7$\%$ increase in PLCC and a 1.5$\%$ increase in SRCC. Moreover, the attribute fusion component yields a 2.2$\%$ increase in PLCC and a 1.8$\%$ increase in SRCC. These findings validate the effectiveness of the Absolute-Attribute Interacting Network in enhancing the fusion of absolute-attribute features and aesthetic features from the image, leading to more substantial improvements. Additionally, modeling relative attributes of images effectively enhances the overall results.

\begin{table*}[!t]
\centering
% \scriptsize
% \tabcolsep=0.11cm
% \renewcommand\arraystretch{1.2}
%\label{tab:att}
\caption{Results of learning partial attributes on the AADB dataset and effect on the TAD66k after branching attributes into the model.}

\begin{tabular}{p{42px}<{\centering}p{42px}<{\centering}p{46px}<{\centering}p{46px}<{\centering}p{46px}<{\centering}p{46px}<{\centering}p{46px}<{\centering}|p{66px}<{\centering}}
\toprule[1.0pt]
Metrics & \makecell{Color \\ Harmony} & \makecell{Shallow \\DoF} & \makecell{Good \\ Lighting} & \makecell{Interesting\\ Content} & \makecell{Rule\\of Thirds} & \makecell{Vivid\\Color} & \makecell{All Attribute\\ on TAD66K} \\ \midrule [0.5pt]
PLCC    & 0.459         & 0.716       & 0.501         & 0.566               & 0.244          & 0.706       & 0.519         \\
SRCC    & 0.453         & 0.492       & 0.433         & 0.564               & 0.236          & 0.699       & 0.494    \\
\bottomrule [1.0pt]
\end{tabular}

\label{tab:att}
% \vspace{-8pt}
\end{table*}

\noindent\textbf{Effectiveness of extracting absolute-attributes with different structures. }
As shown in \cref{tab:attribute structure}, we tested the effectiveness of extracting absolute-attribute features with different structures, mainly focusing on pre-training tasks for attributes such as composition, color, and exposure. We first directly used multiple classic CNN models for testing, and the experimental results demonstrated better extraction of absolute image attributes is beneficial for improving the final aesthetic evaluation. Afterward, we use the combination of features from each layer in InceptionResNetv2 for attribute extraction tasks. The "w/o cnn" and "w/o pool" separately denote that just use the branch with pooling layers and the branch with convolution layers in Absolute-Attribute Perception(AAP). The experimental results demonstrate that AAP which uses both convolution and combined pooling layers simultaneously can fully utilize the extracted features.

\noindent\textbf{Effectiveness of absolute-attribute selection. }
We further test the effects of the selected absolute attributes. Following the approach of other papers \cite{li2023theme, pan2019image}, we use multiple identical Absolute-Attribute Perception Components to learn each absolute attribute of the image from the AADB dataset \cite{aadb}: `Balancing Element', `Color Harmony', `Interesting Content', `Shallow DOF', `Good Lighting', `Rule of Thirds', `Vivid Color'. \cref{tab:att} shows the relatively better results of learning every single attribute on the AADB and the effects of the overall model on the TAD66K dataset after branching each absolute attribute into it. It indicates that although these absolute attributes can effectively promote the aesthetic evaluation task, selecting correct image absolute attributes and using more data to learn the attribute features can achieve better results.

\begin{figure}[t]
   \centering
   \includegraphics[width=1\linewidth]{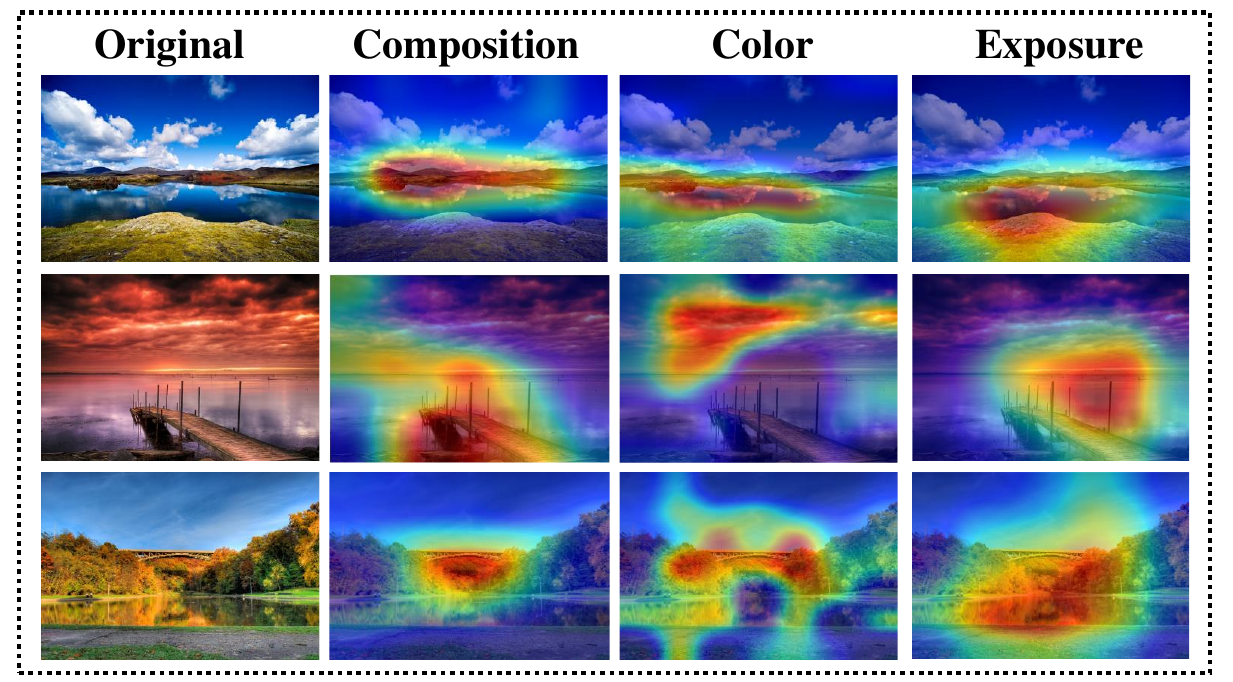}
   % \caption{Qualitative analysis results for each attribute branch. Each attribute branches focus on the regions of the image that can reflect the attribute characteristics}
  \caption{Qualitative analysis results for each attribute branch. It shows that each attribute branches focus on the regions of the image that can reflect the attribute characteristics.}
   \label{fig:cam}
% \vspace{-8pt}
\end{figure}

\subsection{Model Interpretation}
\noindent\textbf{Class Activation Maps. }
We employed Layer Class Activation Mapping (Layer-CAM) \cite{jiang2021layercam} to visualize feature maps for each absolute-attribute branch to highlight model focus areas, as shown in \cref{fig:cam}, while Layer-CAM improves the accuracy of the generated heat map compared to basic CAM. The red areas indicate regions that the model prioritizes. As theme attributes have been extensively studied, our analysis primarily concentrates on composition, color, and exposure attributes.
The visualization reveals that the composition and color branches focus on image regions relevant to their respective attributes. Additionally, the exposure branch emphasizes not only the dark and light areas but also regions related to light. These visualization outcomes affirm the effectiveness of AAP.

\begin{figure*}[t]
\centering
\includegraphics[width=0.98\linewidth]{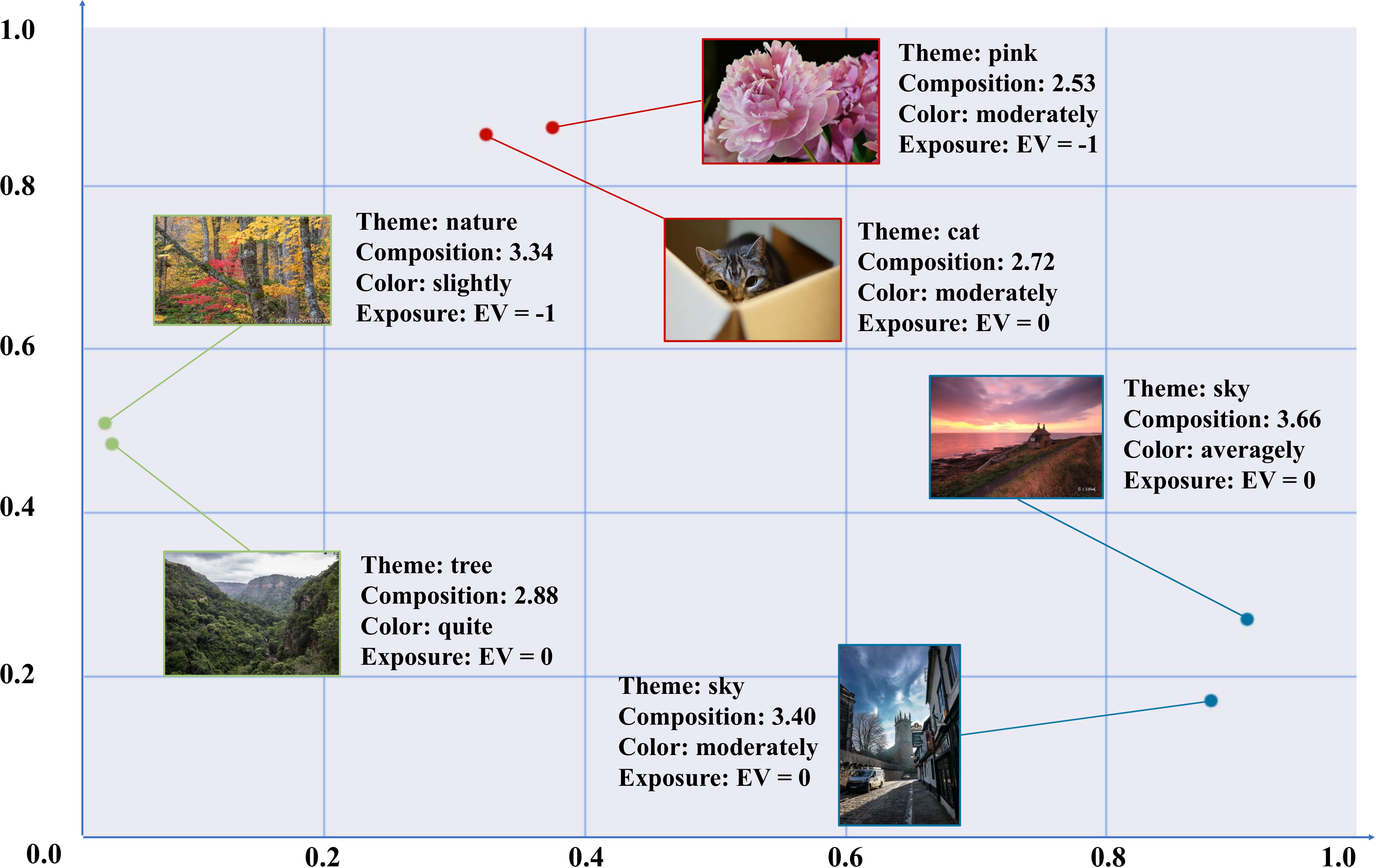}
    \caption{The cluster of Absolute-Attribute Weights after t-SNE transformation of several images on TAD66K test set. The attributes of each image predicted are beside them.}
% \caption{The cluster of Absolute-Attribute Weights after t-SNE transformation of several images on TAD66K test set.}
\label{fig:cluster}
\vspace{-8pt}
\end{figure*}

\begin{figure*}[!t]
   \centering
   \includegraphics[width=0.98\linewidth]{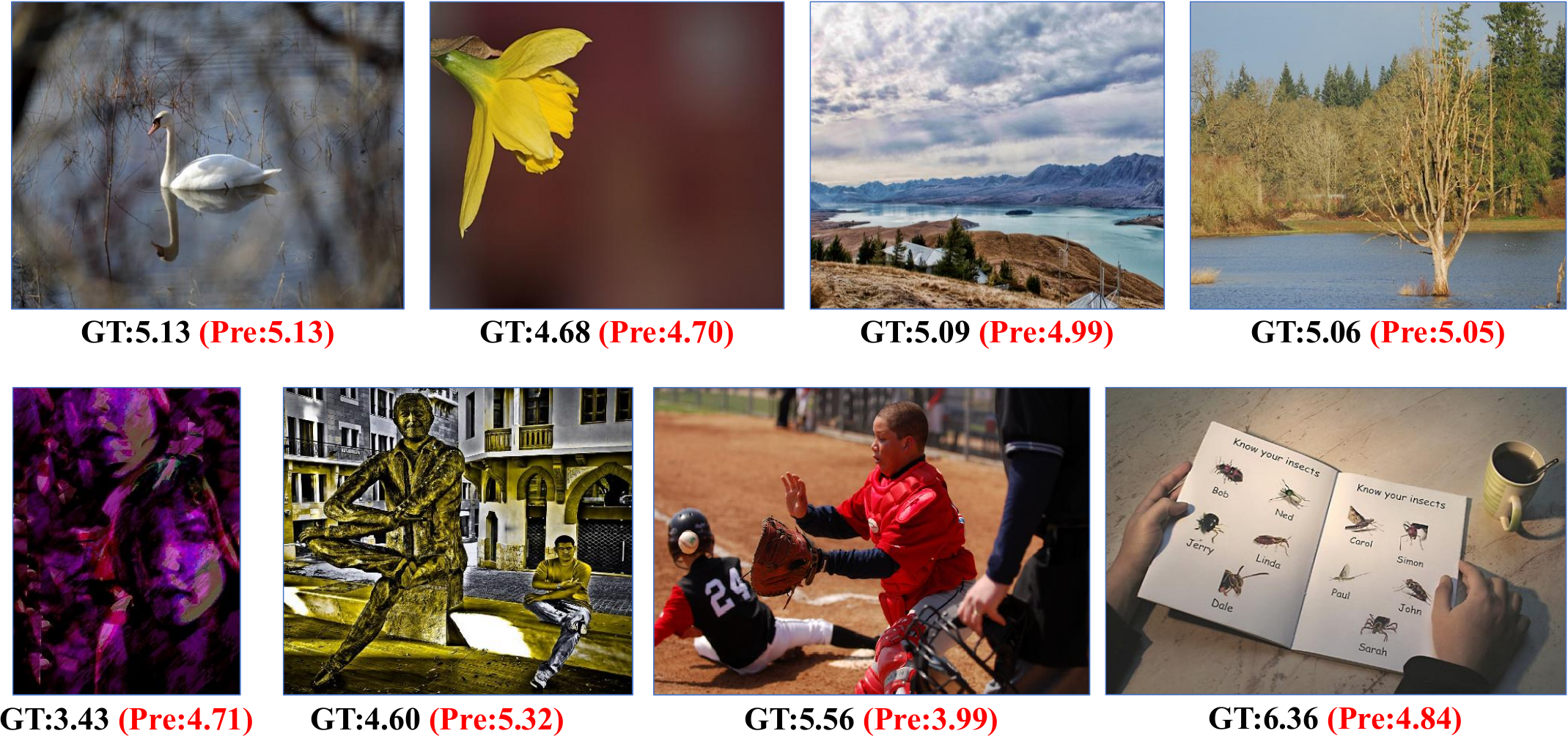}
   % \caption{`GT' and `Pre' means the ground truth and our method's The prediction results respectively. The first row shows several successful cases and the second row shows several failure cases.}
\caption{UMAAF prediction results: successes (first row) and failures (second row). `GT' and `Pre' are the ground truth and prediction results of UMAAF, respecitively.}
   \label{fig:result}
   \vspace{-8pt}
\end{figure*}

\noindent\textbf{Dynamic Fusion of Attribute Features.}
We label each image in the test set by clustering attention weight vectors generated in the Absolute-Attribute Interacting Network. The partial resulting labels are visualized using t-SNE transformation. Each image's weight vector is depicted as a colored dot, and images sharing the same color belong to the same cluster, as seen in \cref{fig:cluster}.
Notably, we observe that images with similar absolute attributes have weight vectors close to each other. For instance, images with blue dots share similar composition, theme, and exposure attributes, with relatively minor differences in color attributes.
Additionally, images with red dots, despite differing themes, display close weight distributions due to similarities in other attributes.
Considering aesthetic feature vectors during weight adjustment, the impact of each image's unique features on overall weighting becomes evident. This leads to similar weighting for feature vectors in images with distinct attributes, as seen in green-framed images. 
This phenomenon further validates the self-adaptive nature of the module.

\noindent\textbf{Prediction Results Analysis. }
We further analyze the success and failure prediction outcomes of UMAAF, as shown in \cref{fig:result}. The first row shows several successful cases and the second row shows several failure cases. In the first row, it shows that if the image has sufficiently distinct and explicit attributes, our predictions tend to be better. And if the aesthetic quality of the image is closely related to its semantic content, as shown in the second row, UMAAF understands these abstract semantics inadequately, resulting in significant prediction errors. 

\section{Conclusions}
\label{sec:conclusion}
In this paper, we introduce UMAAF, a comprehensive framework that handles both absolute and relative attributes of images for IAA. 
We extract determined absolute attributes by the Absolute-Attribute Perception Component and propose an Absolute-Attribute Interacting Network that dynamically learns attribute weights, effectively integrating diverse absolute-attribute perspectives and generating aesthetic predictions.
For modeling relative attributes, we introduce the Relative-Relation Loss, considering the relative rankings and distance relationships between images, further enhancing performance. Extensive experiments demonstrate our model's state-of-the-art performance and its alignment with human preference.
However, our approach still has limitations. There is still a significant unexplored territory in attribute selection and utilization for image aesthetic assessment. In future work, we intend to conduct further in-depth research in related areas to advance image aesthetic understanding.

%\newpage

\bibliographystyle{IEEEbib}
\bibliography{total}

\newpage
\clearpage
\setcounter{page}{1}
\maketitlesupplementary

\appendix

\setcounter{figure}{0}
\renewcommand*{\thefigure}{A\arabic{figure}}
\setcounter{table}{0}
\renewcommand*{\thetable}{A\arabic{table}}

% \clearpage
% \setcounter{page}{1}
% \maketitlesupplementary

This supplementary material will provide more description, experiment, and visualization results, 
% and they are organized as follows:
and organized as outlined below:

\begin{itemize}
\item Section \ref{section:a}: Extended samples and detailed insights into the absolute-attribute pre-training task.
This section presents further samples and elaborates on the nuances of our absolute-attribute pre-training task.

\item Section \ref{section:b}: In-depth architecture of UMAAF and encompassed features.
Detailed information is provided regarding the architecture of UMAAF and the various features it encompasses.

\item Section \ref{section:c}: Comprehensive experiments on the balancing coefficient of the Relative-Relation Loss and its derivation.
This section expands on the experimentation involving the balancing coefficient of the Relative-Relation Loss. Furthermore, it offers a more detailed derivation process for this coefficient.

\item Section \ref{section:d}: Elaborate clarification of the dataset and utilized evaluation metrics.
A thorough explanation of the dataset utilized in our study is provided in this section, along with a detailed overview of the evaluation metrics we have adopted.

\item Section \ref{section:e}: Extended visualized results and in-depth analysis.
This section presents a broader range of visualized outcomes and delves deeper into the analysis of these results.

\item Section \ref{section:f}: We will present more details on the ablation experiment of AAP.

\item Section \ref{section:g}: Some examples of the impact of changing image attributes on image aesthetic scores.

\end{itemize}

% Sec \ref{section:b} shows more samples and details about our absolute-attribute pre-training task.

% Sec \ref{section:c} provides more details of the architecture of the UMAAF and the features in it.

% Sec \ref{section:d} conducts more experiments about the balancing coefficient of the Relative-Relation Loss and provide more derivation process of it.

% Sec \ref{section:e} provides a detailed explanation of the dataset and evaluation metrics we have adopted.

% Sec \ref{section:f} gives more visualized results and analysis.

% \section{Problem Definition}
% The essential peculiarity of few-shot anomaly detection (FSAD) is that there are only a few normal samples in the training set.
% Given a training set $S_{train} = \{S_{train}^{1}, S_{train}^{2}, ... , S_{train}^{n}\}$, 
%  which consists of only normal images from n categories. 
% The testing set has normal or abnormal images from each category $S_{test} = {S_{test}^{1}, S_{test}^{2}, ... , S_{test}^{n}}$
% We follow the one-class strategy and focus on only one category at a time without training the model.
% For a target category $S_t$, available samples $I_i$ $( i\in \{ 1, 2, ..., n\})$ are normal and generally $n\leq8$.
% On the above basis, FSAD requires the model to identify whether the specific image has anomalies and localize the anomalous pixels.

\appendix

% \section{Overview}

\section{More details of pre-training tasks}
\label{section:a}
\begin{figure}[t]
   \centering
   \includegraphics[width=1\linewidth]{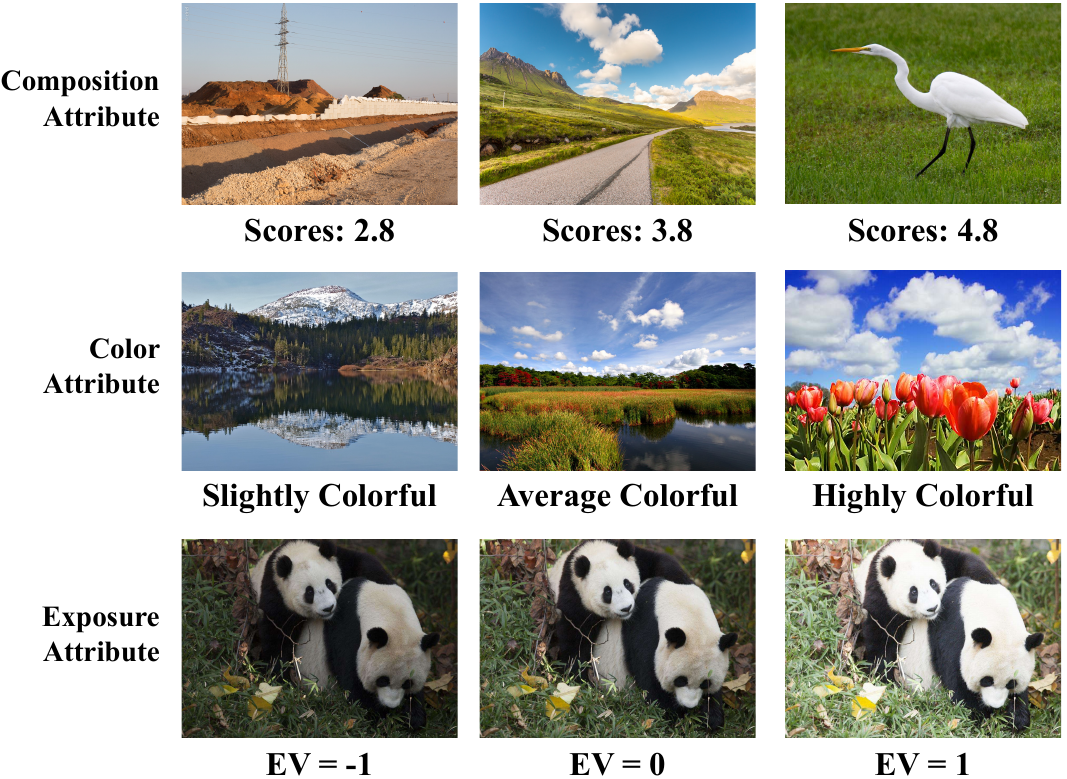}
   % \caption{The absolute-attribute weights of several images in test set of TAD66K are clustered and then plotted in the 2D space via t-SNE transformation. The attributes of each images predicted by Attribute Perception Module are beside them. The figure shows that the weight feature vectors of the images with different absolute-attribute cases have different distribution.}
    \caption{Some samples of three pre-training tasks.}
   \label{fig:samples}
\end{figure}

We use Absolute-Attribute Perception Components to pre-train on datasets corresponding to three absolute attributes: composition, color, and exposure. Some samples of three datasets are shown in \cref{fig:samples}. The details regarding the composition attribute pre-training task have already been explained in the main paper and for the color attribute, we get a $90.82\%$ accuracy on the test set.

For the exposure attribute, we employ the dataset from \cite{afifi2021learning} for pre-training. In order to facilitate better pre-training, we input correctly exposed images in the dataset along with the images that need to be classified. Due to the goal of directing the model's attention to specific regions in the images rather than achieving high classification scores, we use all images in the dataset to pre-train and get a $89.84\%$ accuracy on the train set.

We adopt Mean Squared Error(MSE) loss for the pre-training task of composition attribute and Cross Entropy loss for the pre-training tasks of color and exposure attributes.

% \begin{figure*}[h]
%   \centering
%     \begin{subfigure}
%       \centering   
%       \includegraphics[width=0.82\linewidth]{./figs/com.pdf}
%     \caption{More visualization results about composition attribute.}
%         \label{fig:com}
%     \end{subfigure}   %      \hfill  % 这个\hfill指令为插入弹性长度的空白，看情况选择加不加。
%     \begin{subfigure}
%       \centering   
%      \includegraphics[width=0.82\linewidth]{./figs/col.pdf}
%     \caption{More visualization results about color attribute.}
%         \label{fig:col}
%     \end{subfigure}
%   \begin{subfigure}
%   \centering   
%  \includegraphics[width=0.82\linewidth]{./figs/exp.pdf}
% \caption{More visualization results about exposure attribute. }
%     \label{fig:exp}
% \end{subfigure}
% % \caption{
% % \label{fig:total}
% % write\_caption\_here
% % }
% \end{figure*}

\begin{figure*}[t]
   \centering
   \includegraphics[width=1\linewidth]{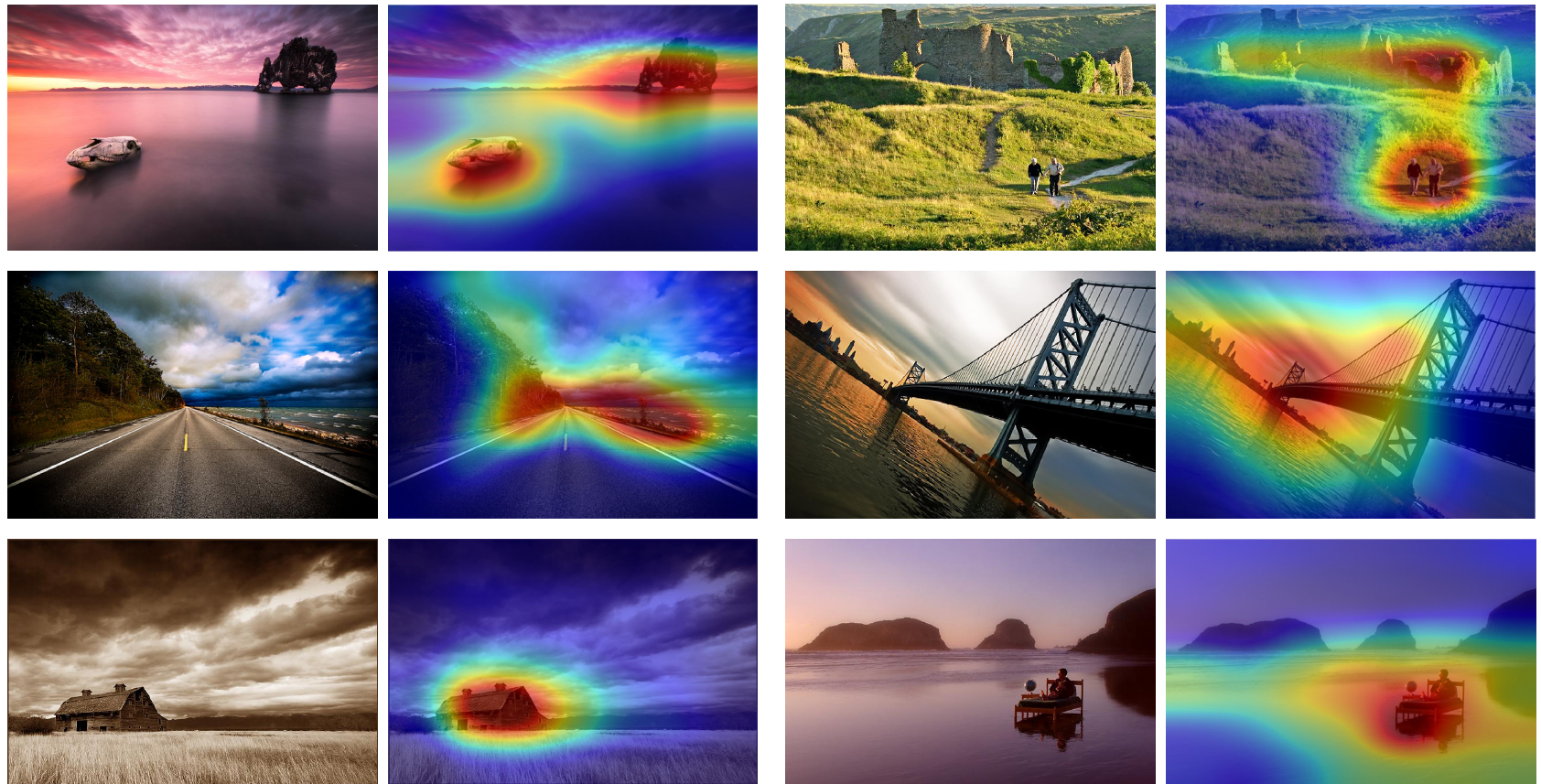}
    \caption{More visualization results about composition attribute. }
   \label{fig:com}
\end{figure*}

% \begin{figure*}
%   \centering
%     \begin{subfigure}
%       \centering   
%      \includegraphics[width=0.92\linewidth]{./figs/com.pdf}
%     \caption{More visualization results about composition attribute.}
%         \label{fig:col}
%     \end{subfigure}
%   \begin{subfigure}
%   \centering   
%  \includegraphics[width=0.92\linewidth]{./figs/exp.pdf}
% \caption{More visualization results about color attribute. }
%     \label{fig:exp}
% \end{subfigure}
% \end{figure*}

\section{More details of the UMAAF}
\label{section:b}
In Absolute-Attribute Perception Component, we first resize the feature maps via `area' interpolate and concatenate the feature maps from multi Conv blocks into one feature map with the size of $16928\times 7 \times 7$. Then, through the next two parts of the Absolute-Attribute Perception Component, we get two feature maps with the size of $1024\times 7 \times 7$ respectively and concatenate them into one feature map with the size of $2048\times 7 \times 7$, which is sent to Absolute-Attribute Interacting Network.
The feature maps from the theme extractor and Aesthetics Perceiving Network are resized into $512\times7\times 7$ and $1280 \times 7 \times 7$ respectively.

In Absolute-Attribute Interacting Network, all feature maps sent in are resized into 
$512 \times 7 \times 7$ through multiple conv layers with $1\times1$ kernel and then, they are concatenated and sent to channel attention block. In the channel attention block, the concatenation of feature maps is resized into two feature vectors through an Average Pooling Layer and Max Pooling Layer. Then we use them to calculate two channel attention vectors through one MLP and get the final channel attention vectors by element-wise summation. The process of the channel attention block can be formulated as follows:

\begin{equation}
\begin{aligned}
h=\sigma(MLP(f_1(c))+MLP(f_2(c))) \odot c,
\end{aligned}
\end{equation}
where c denotes the concatenation of absolute-attribute features and generic aesthetic feature before the channel attention block, and $f_1$ and $f_2$ denote the Max Pooling and Average Pooling. The MLP used is the same.

After that, we get out the absolute-attribute features and convert them into feature vectors with the size of $512\times 1 \times 1$ via an Average Pooling Layer. 

In the final feature fusion, we use the bilinear fusion \cite{Mao_Zhu_Su_Cai_Li_Dong_2023} to fuse the overall attribute feature and generic aesthetic feature, its formulation can be set as follows:

\begin{equation}
\begin{aligned}
\hat{p} &= W_1y_1 + W_2y_2 + y^T_1W_3y_2 + b,
\end{aligned}
\end{equation}
where $y_1$ and $y_2$ denote the concatenation of all absolute attribute feature vectors and generic aesthetics feature vector respectively. W1, W2, and W3 denote the learnable parameters and
b is a real number. $\hat{p}$ represents the output of the model.

\section{More details and experiments of Relative-Relation Loss}
\label{section:c}
The introduced Relative-Relation Loss is implemented on the triplet loss \cite{fang2021superpixel}. Referring to the derivation in \cite{9706735},
in the condition of $g_i>g_j>g_k$,  we want $|p_i-p_j|-|p_i-p_k|+margin\leq0$, and since in a hypothetically perfect prediction environment, the predicted results can be set as the ground truth, so, the inequality can be converted into $g_k-g_j+margin\leq0$, thus the margin can be set as the upper limit $g_j-g_k$. By the same process, the margin can be set as $g_k-g_j$ in the condition of $g_i<g_j<g_k$.
Thus, the margin of the used triplet loss is set as  $|g_j-g_k|$.

For the balancing coefficient $\lambda$ of the Relative-Relation Loss, we conduct more experiments about it on TAD66K dataset, and their results are shown in \cref{tab:bc}.

\begin{table}[h]
%\caption{Comparison of the proposed model with several representative IAA methods on human preference.}
\caption{Results with different balancing coefficient of Relative-Relation Loss. }
\tabcolsep=0.47cm
\renewcommand\arraystretch{1.2}
%\label{tab:human}
\begin{tabular}{ccc}
\toprule[1.0pt]
% \usepackage{booktabs}
Balancing Coefficient $\lambda$                      & PLCC $\uparrow$  & SRCC $\uparrow$ \\ \midrule[1pt]
0.01                          &0.537  &0.512 \\
\textbf{0.05}                      & \textbf{0.540}  & \textbf{0.515} \\
0.1                        & 0.534  & 0.509 \\
0.5                     & 0.528  & 0.506 \\ \bottomrule[1.0pt]
\end{tabular}
%\caption{Results with different balancing coefficient of Relative-Relation Loss. }
\label{tab:bc}
% \vspace{-8pt}
\end{table}

\begin{figure*}[!h]
   \centering
   \includegraphics[width=1\linewidth]{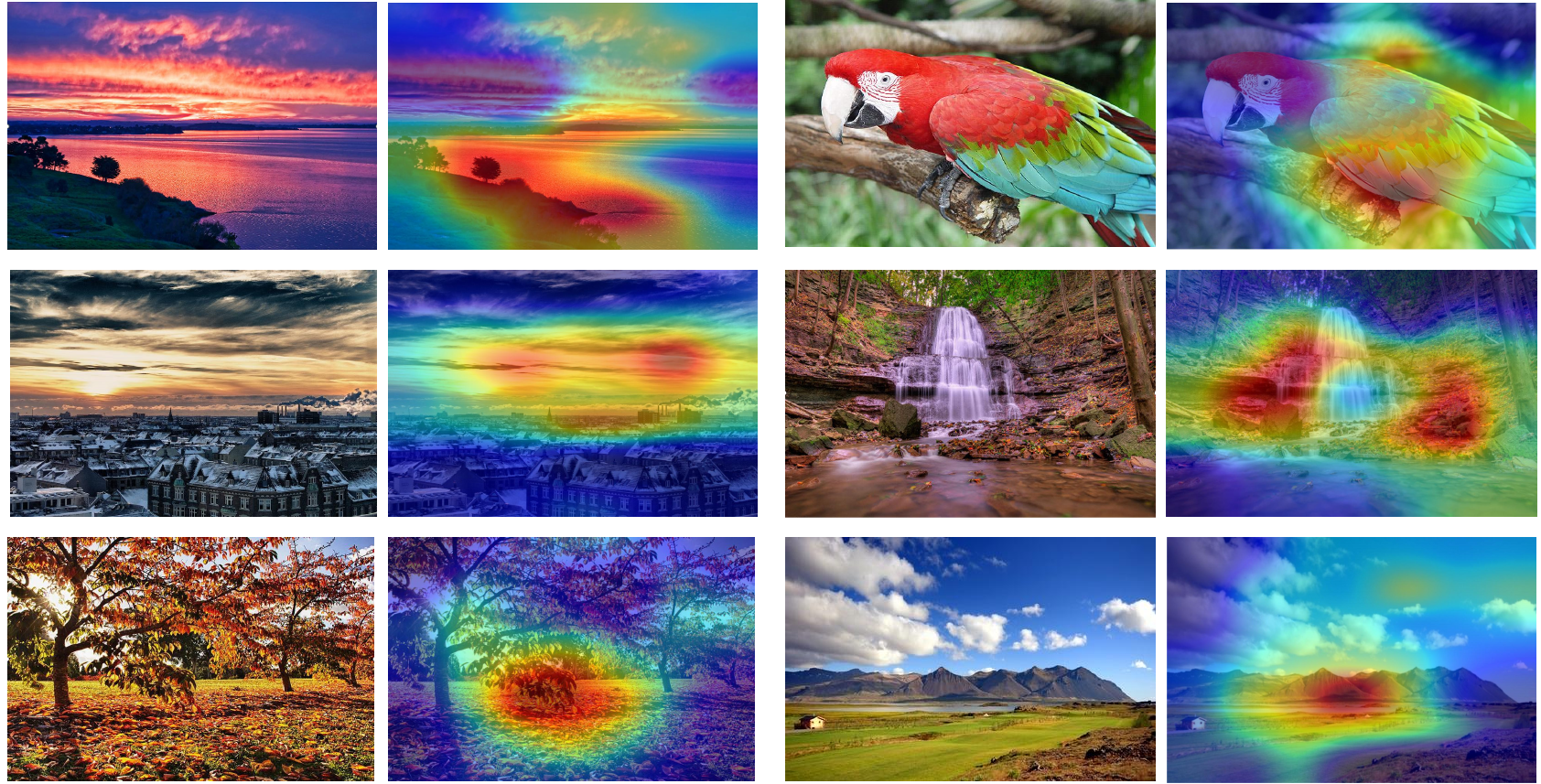}
    \caption{More visualization results about color attribute.}
   \label{fig:col}
\end{figure*}

\begin{figure*}[!h]
   \centering
   \includegraphics[width=1\linewidth]{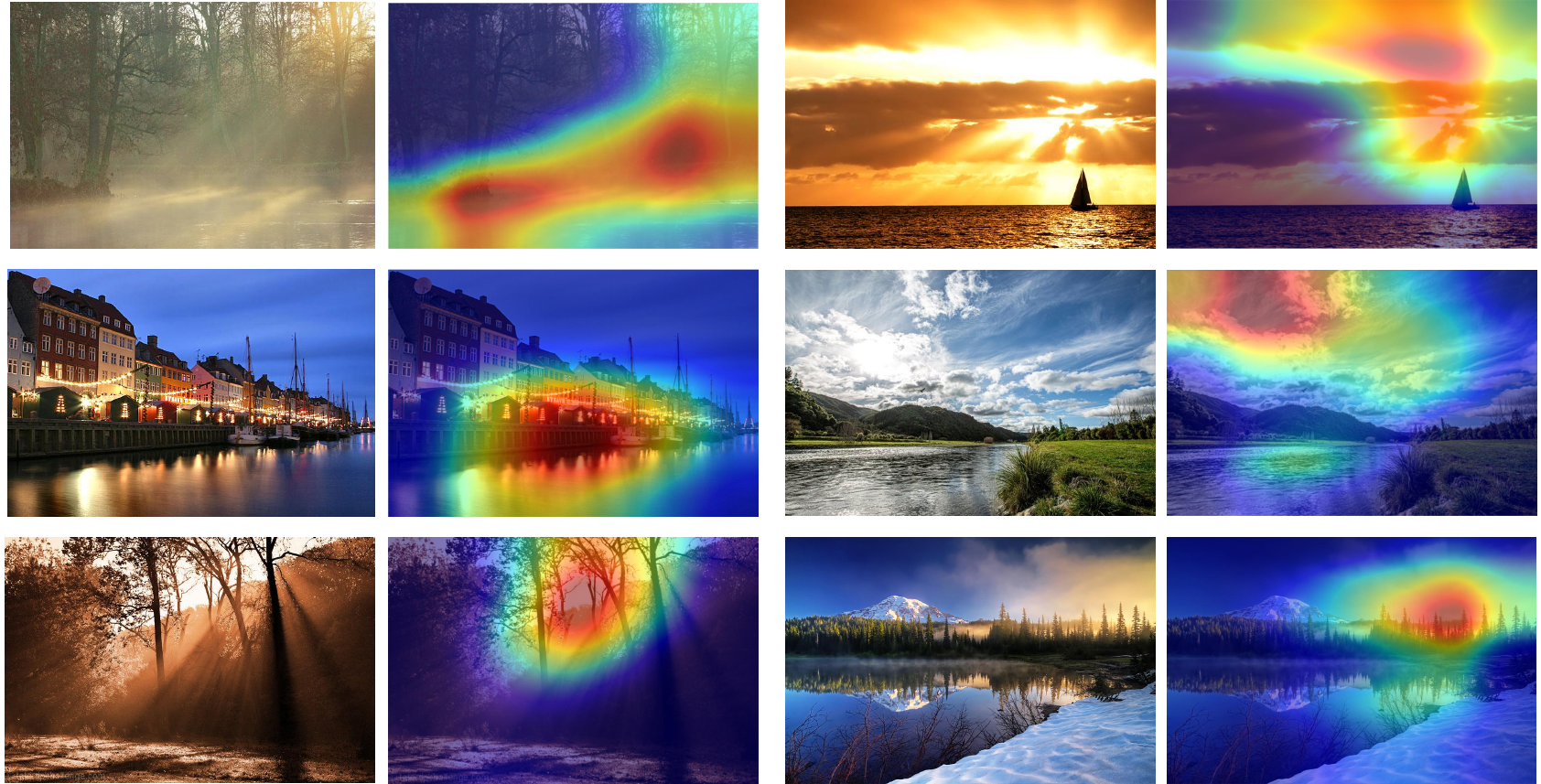}
    \caption{More visualization results about exposure attribute.}
   \label{fig:exp}
\end{figure*}

\section{Datasets and Evaluation Metrics}
\label{section:d}
\noindent\textbf{Datasets. }
Our experiments are mainly carried out on TAD66K \cite{he2022rethinking} and AVA \cite{6247954}.

TAD66K dataset\cite{he2022rethinking}  is a newly proposed aesthetic assessment dataset and contains 66327 images with at least 1200 valid annotations for each image, which is more than any other aesthetic assessment dataset at present. Moreover, measures are taken to effectively alleviate the problem of long-tailed distribution that exists in other aesthetic evaluation datasets. Since the accuracy and universality of the labels, we primarily conduct further experiments on the TAD66k dataset.

AVA dataset\cite{6247954} contains more than 255,000 images, each of which is voted on by 78-549 viewers with a voting score ranging from 1 to 10. The data partitioning is set up as in the previous work \cite{6247954,he2022rethinking}. The score threshold is set as 5.0, classifying images with an average aesthetic score above 5.0 as high aesthetic quality and images with an average aesthetic score below 5.0 as low aesthetic quality.

\vspace{10pt}
\noindent\textbf{Evaluation Metrics. }
In the evaluation phase, Pearson Linear Correlation Coefficient (PLCC) and Spearman’s Rank Correlation Coefficient (SRCC) are used to evaluate the image scores predicted by the model. They are metrics used to measure the correlation between predicted scores and ground truth. The higher they are, the better the model performs. Specifically, an accuracy metric is used to evaluate the models' ability to classify high and low aesthetic quality on the AVA dataset.
In addition, the MSE loss and EMD loss on test sets are also included in the evaluation metrics on TAD66K and AVA respectively to assess the model's performance.

\begin{figure}[!h]
   \centering
   \includegraphics[width=1\linewidth]{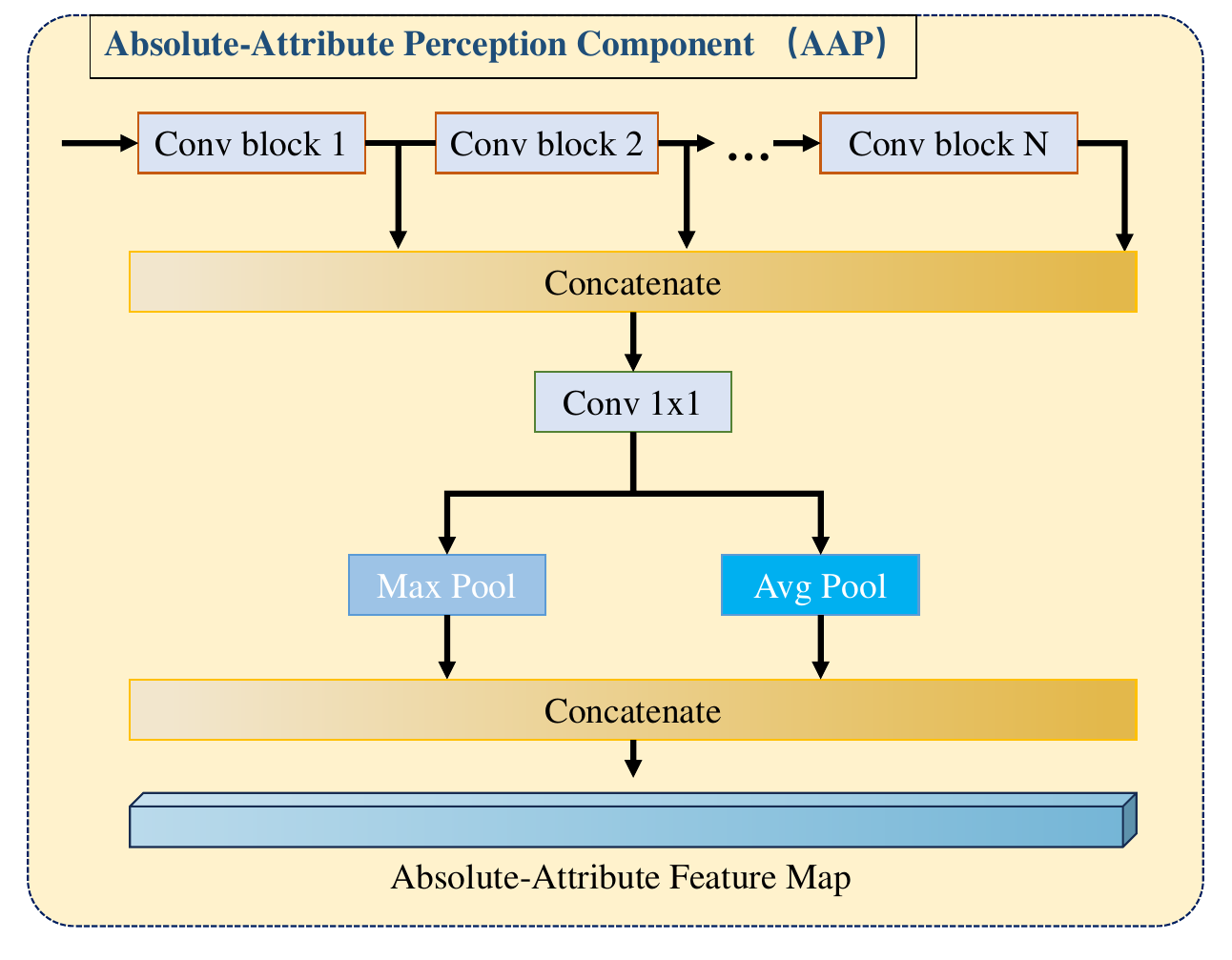}
    \caption{Corresponding AAP structure for `w/o cnn' setting.}
   \label{fig:wocnn}
\end{figure}

\begin{figure}[!h]
   \centering
   \includegraphics[width=1\linewidth]{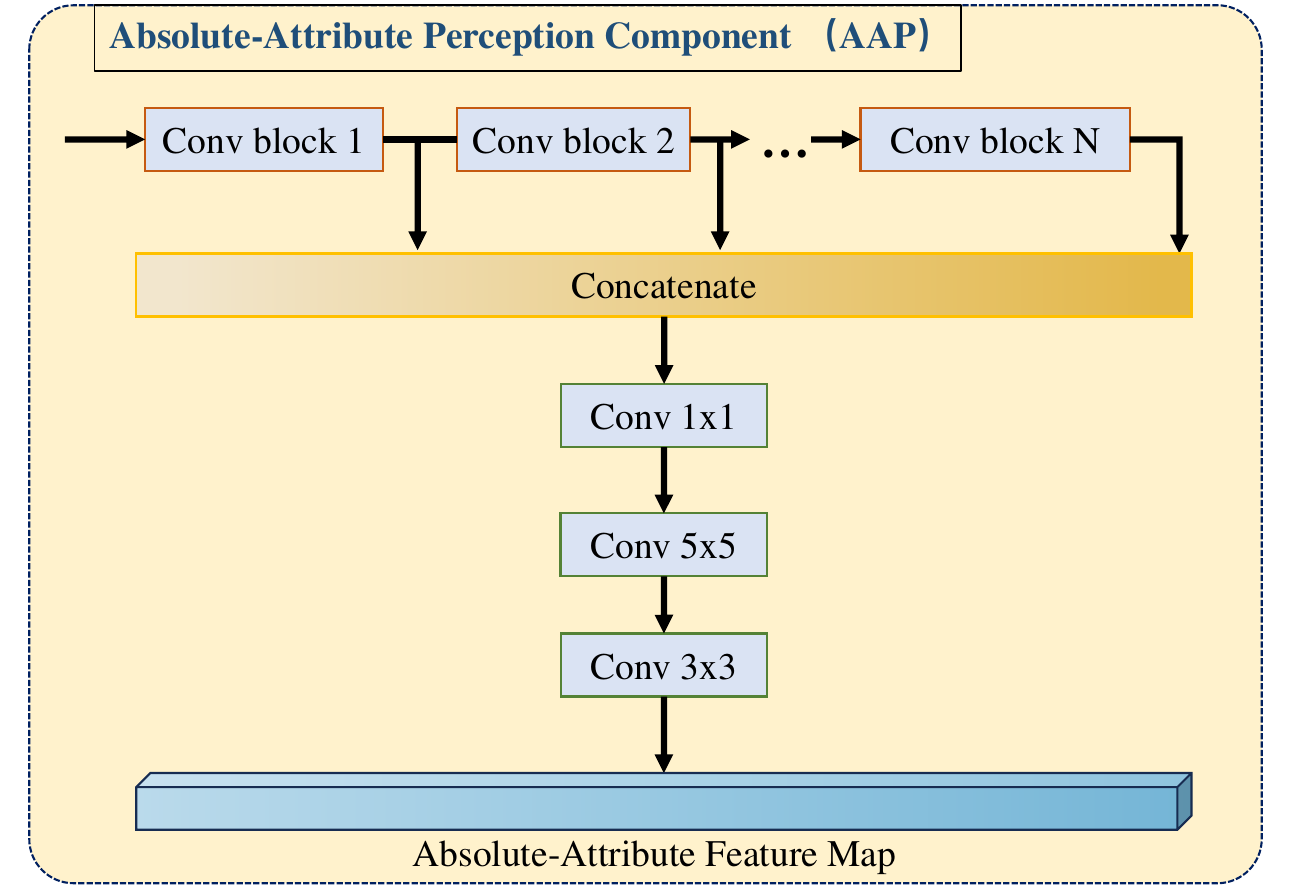}
    \caption{Corresponding AAP structure for `w/o pool' setting.}
   \label{fig:wopool}
\end{figure}

\begin{figure}[!h]
   \centering
   \includegraphics[width=1\linewidth]{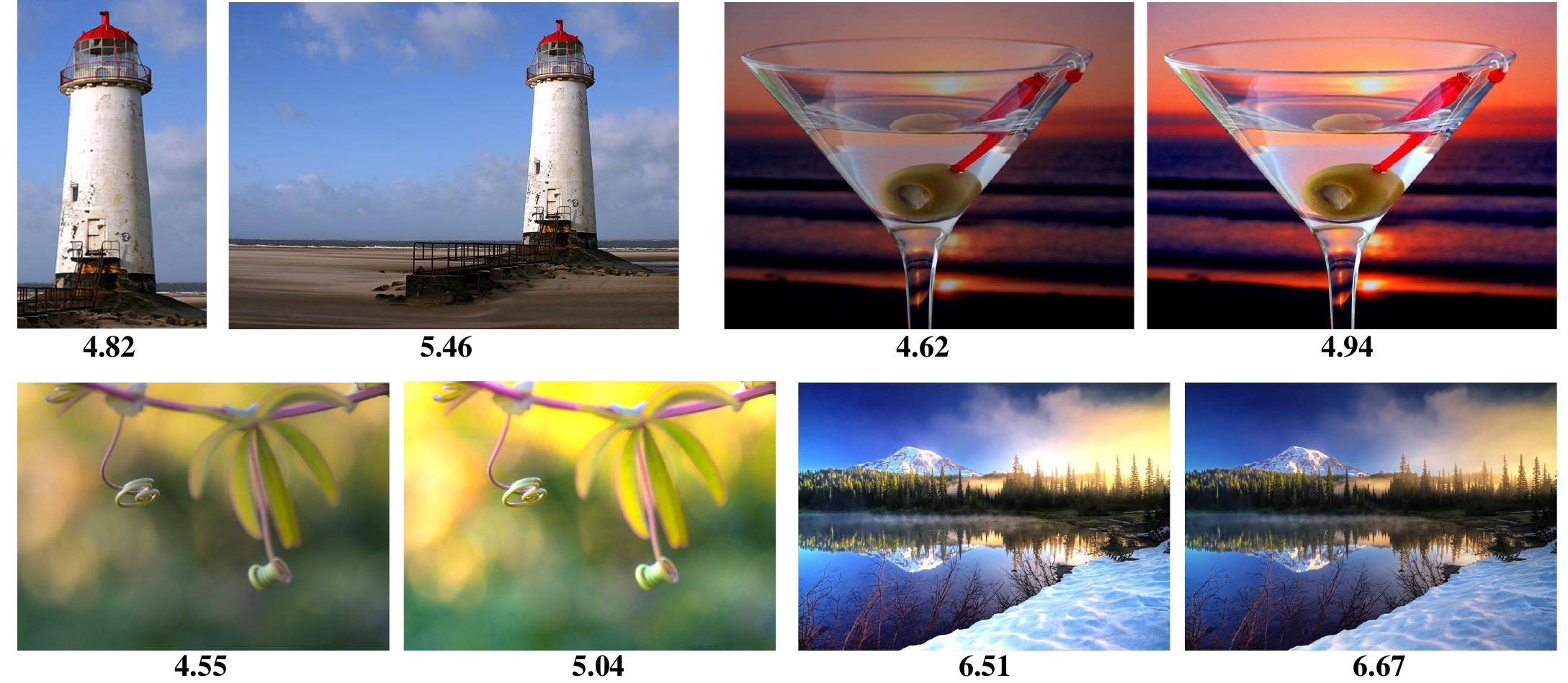}
    \caption{Examples of changing the aesthetics attributes of the images. The scores below each images are their aesthetics scores predicting by our model.}
   \label{fig:appendixatt}
\end{figure}

\section{More Visualization Results}
\label{section:e}
As shown in \cref{fig:com}, \cref{fig:col}, \cref{fig:exp}, we show more comprehensive visualization results about different absolute-attributes extractors.
In \cref{fig:com}, when faced with more diverse composition types, our composition attribute extractor is still able to focus on the areas that influence the composition of the image.
In \cref{fig:col}, our color attribute extractor focuses on colors in the image as much as possible, thus extracting more comprehensive color information.
In \cref{fig:exp}, it can be seen that the areas about light information which will significantly influence the image's exposure level and also has a significant impact on the aesthetics of the image have also received more attention.

\section{More details of the ablation study of AAP}
\label{section:f}
In the main text, we present the ablation experimental results of AAP with different structures. Among them, `w/o cnn' and `w/o pool' respectively represent two situations where pool layer and cnn layer are only applied on the connected feature map, and their specific structures are shown in \cref{fig:wocnn} and \cref{fig:wopool}.

\section{Some examples of the change of aesthetics scores after changing aesthetics attributes.}
\label{section:g}
In this section, we attempt to change some aesthetic properties of the image, such as image brightness, color saturation, and composition. As shown in \cref{fig:appendixatt}, reasonably modifying the aesthetic attributes of an image can effectively improve its visual appeal. It also indicates that a good aesthetic evaluation model can guide the image modification in the real life.

% \begin{figure*}[h]
%   \centering
%     \begin{subfigure}
%       \centering   
%      \includegraphics[width=1\linewidth]{./figs/col.pdf}
%     \caption{More visualization results about color attribute.}
%         \label{fig:col}
%     \end{subfigure}
%   \begin{subfigure}
%   \centering   
%  \includegraphics[width=1\linewidth]{./figs/exp.pdf}
% \caption{More visualization results about exposure attribute. }
%     \label{fig:exp}
% \end{subfigure}
% \end{figure*}

\end{document}